\documentclass[runningheads]{llncs}
\usepackage[T1]{fontenc}
\usepackage{graphicx,verbatim}
\usepackage{amsmath}
\usepackage{amssymb}
\usepackage{amsmath}
\usepackage{booktabs}
\usepackage{hyperref}
\usepackage[table,xcdraw]{xcolor}

\usepackage[misc]{ifsym}

\definecolor{BDE6CD}{HTML}{BDE6CD}
\definecolor{E4EEBC}{HTML}{E4EEBC}
\definecolor{FFF8C5}{HTML}{FFF8C5}

\begin{document}
\titlerunning{MFM-DA: IA Adaptor and Hierarchical Alignment for EDA in MFMs}
\title{MFM-DA: Instance-Aware Adaptor and Hierarchical Alignment for Efficient Domain Adaptation in Medical Foundation Models}
%

\author{Jia-Xuan Jiang\inst{1,\dagger} \and Wenhui Lei\inst{2,3,\dagger} \and Yifeng Wu\inst{4} \and Hongtao Wu\inst{5} \and Furong Li\inst{1} \and Yining Xie\inst{1} \and Xiaofan Zhang\inst{2,3,}$^{\textrm{\Letter}}$ \and Zhong Wang\inst{1,}$^{\textrm{\Letter}}$}
\authorrunning{J. Jiang and W. Lei et al.}
\institute{Lanzhou University \and Shanghai Jiaotong University \and Shanghai Artificial Intelligence Laboratory  \and Shenzhen Institute of Advanced Technology, Chinese Academy of Sciences \and The Hong Kong University of Science and Technology (Guangzhou) \\
    \email{xiaofan.zhang@sjtu.edu.cn, wangzhong@lzu.edu.cn}}

\renewcommand{\thefootnote}{\fnsymbol{footnote}}
\footnotetext{$\dagger$ Contributed equally to this work.}
    
\maketitle              
\begin{abstract}
Medical Foundation Models (MFMs), trained on large-scale datasets, have demonstrated superior performance across various tasks. However, these models still struggle with domain gaps in practical applications. Specifically, even after fine-tuning on source-domain data, task-adapted foundation models often perform poorly in the target domain. To address this challenge, we propose a few-shot unsupervised domain adaptation (UDA) framework for MFMs, named MFM-DA, which only leverages a limited number of unlabeled target-domain images. Our approach begins by training a Denoising Diffusion Probabilistic Model (DDPM), which is then adapted to the target domain using a proposed dynamic instance-aware adaptor and a distribution direction loss, enabling the DDPM to translate source-domain images into the target domain style. The adapted images are subsequently processed through the MFM, where we introduce a designed channel-spatial alignment Low-Rank Adaptation (LoRA) to ensure effective feature alignment. Extensive experiments on optic cup and disc segmentation tasks demonstrate that MFM-DA outperforms state-of-the-art methods. Our work provides a practical solution to the domain gap issue in real-world MFM deployment. Code will be available at \href{https://github.com/HopkinsKwong/MFM-DA}{here}.

\end{abstract}
\section{Introduction}

Deep learning has made significant strides in medical image analysis \cite{MambaSAM,PCSDG,lei2021automatic}. However, these models often require large volumes of annotated data for training and tend to be task-specific, which limits their generalization across diverse clinical applications. To overcome this limitation, various MFMs have recently been proposed \cite{zhang2024challenges,UNI,zhou2023foundation,lei2025medlsam,lei2025data}, which learn universal representations from vast medical datasets. These models are capable of performing a wide range of clinical tasks either directly or after fine-tuning.

\begin{figure}[t]
	\centering
	\includegraphics[width=0.8\textwidth]{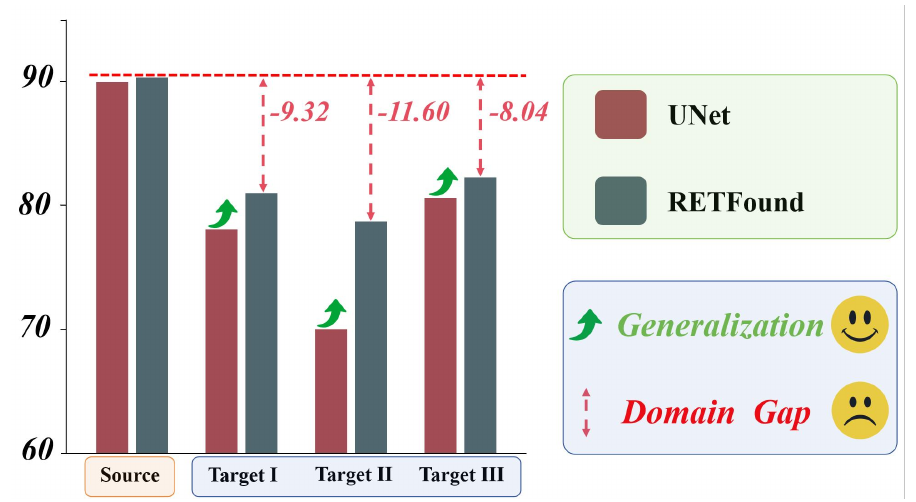}
	\caption{Performance comparison of RETFound on the cup-to-disc segmentation task for fundus images. Despite showing superior generalization compared to UNet trained from scratch, RETFound still suffers from domain shifts, with a noticeable decline in performance when applied to the target domain.}
	\label{intro}
\end{figure}

When fine-tuned on specific tasks within a source domain, these MFMs exhibit superior generalization performance in the target domain compared to models trained from scratch. However, they remain susceptible to domain shifts, often showing a noticeable decline in performance when applied to the target domain. For example, RETFound\cite{zhou2023foundation}, a foundation model trained on 1.6 million unannotated retinal images using self-supervised learning, was evaluated on cup-to-disc segmentation tasks for fundus images. Our experiments reveal that RETFound demonstrates superior generalization compared to models trained from scratch, but its performance still suffers from domain gaps, as shown in Fig. \ref{intro}. Despite outperforming UNet \cite{ronneberger2015u} trained from scratch in the target domain, RETFound's results are significantly lower than its performance in the source domain, highlighting the impact of domain shifts.

To mitigate domain shifts, unsupervised domain adaptation (UDA) \cite{ganin2015unsupervised} has been widely explored to improve model performance on unannotated target data using labeled source domain data. UDA typically addresses domain shifts in two ways: image adaptation \cite{DAimage}, which aligns the image appearances through pixel-level transformations, and feature adaptation \cite{Coral}. However, UDA often requires large amounts of unannotated target data to reduce domain distribution differences, which is not always feasible in real-world medical scenarios. In contrast, Few-shot Domain Adaptation (FSDA) \cite{motiian2017few} offers a more practical solution, as it only requires a limited number of target samples during training.

To address the challenge of domain shifts in MFM with limited target data, we propose the MFM-DA framework, which requires only source domain data and a small number of target-domain images. As shown in Fig. \ref{model}, it mainly contains two stages. In Stage 1, we train a Denoising Diffusion Probabilistic Model (DDPM) \cite{DDPMLoss} on source domain data, which is then adapted to the target domain using our proposed Dynamic Instance-Aware Adaptator and distribution consistency loss. In Stage 2, we fine-tune the foundation model with a combination of source and generated target-domain images, using LoRA \cite{Rein} for adjusting attention mechanisms and a Pyramid Hierarchical Alignment method to align features across hierarchical levels. This approach facilitates domain adaptation by ensuring alignment in both channel-wise semantics and spatial structures, thereby improving model performance in the target domain.

We conducted extensive experiments on optic cup and disc segmentation tasks, demonstrating the effectiveness of our approach across source and target domains for MFMs. Our contributions include: \begin{enumerate} \item To our best knowledge, we are the first to propose a specifically designed framework to address the few-shot unsupervised domain adaptation for medical foundation models; \item Introducing the Dynamic Instance-Aware Adaptator, which adapts the distribution of generated images to better match the target domain, even in few-shot scenarios. \item Proposing Pyramid Hierarchical Alignment, which aligns source and target-domain features at different levels to achieve robust domain adaptation. 
\end{enumerate}

\section{Method}
\begin{figure}[t]
	\centering
	\includegraphics[width=1\textwidth]{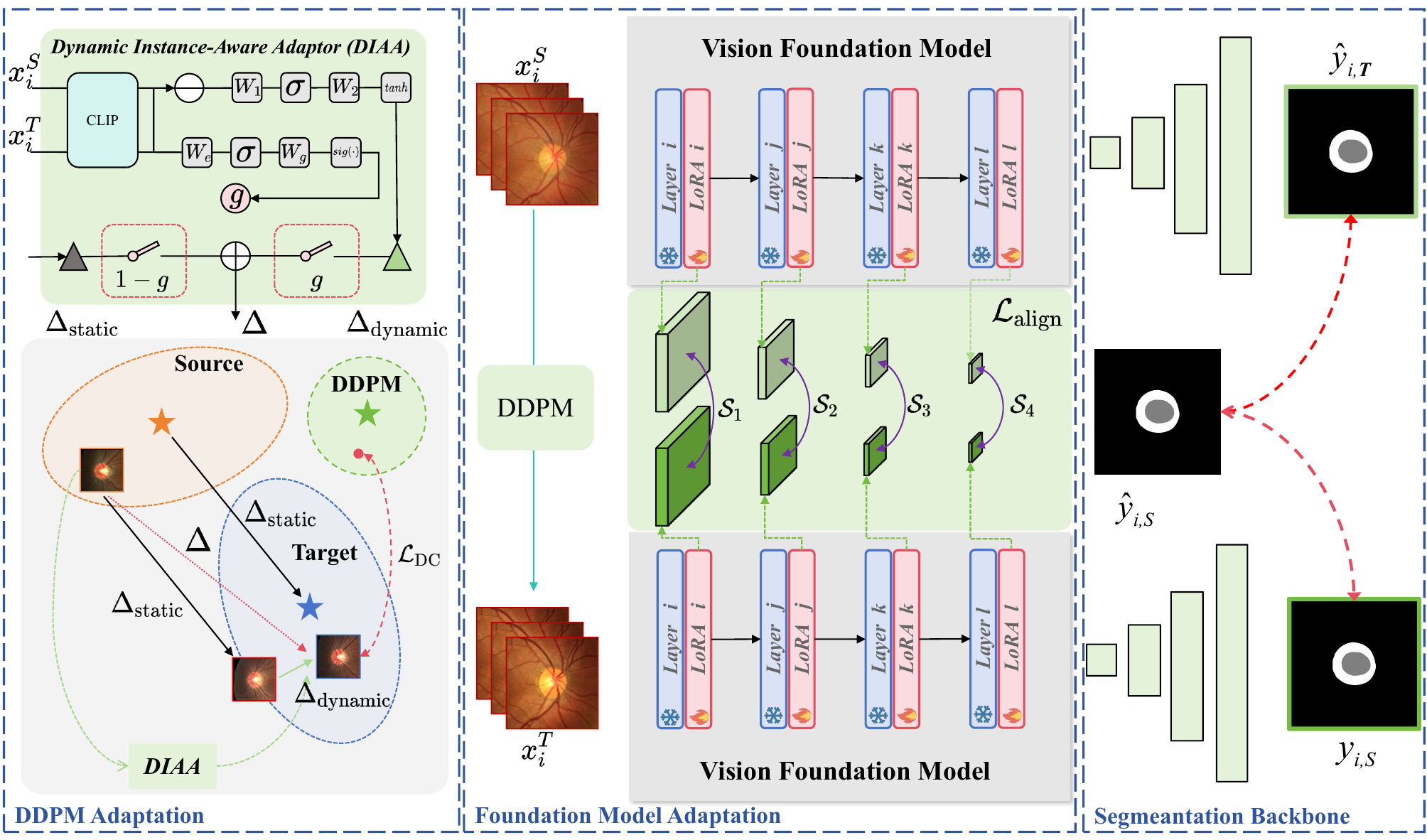}
	\caption{The proposed MFM-DA framework aims to perform domain adaptation of the foundation model by enabling the DDPM to adapt to target domain images in a few-shot setting. This process generates a large number of target domain images, which are then utilized for domain adaptation in the foundation model. The framework includes an Instance-aware Adaptor, which dynamically adjusts the adaptation direction of DDPM in the feature space, and a Hierarchical Alignment loss that aligns the pyramid features of the Foundation Model.}
	\label{model}
\end{figure}

The framework we propose for few-shot unsupervised domain adaptation in medical image segmentation is illustrated in Fig. \ref{model}. We introduce a two-stage domain adaptation framework. In the first stage, DDPM is used to perform domain adaptation on 10 unannotated target domain images from the perspective of image generation distribution. In the second stage, we fine-tune the foundation model on the generated unannotated target domain images to perform domain adaptation at the feature level. This approach effectively narrows the performance gap caused by domain differences in the foundation model.

\subsection{Dynamic Instance-Aware Adaptor}
In the few-shot scenario, models are highly susceptible to overfitting. To address the overfitting issue, \cite{DDC} proposes the Directional Distribution Consistency loss (DDC), which to extracts the feature distribution centers of all data from both the source domain and the target domain, and relies on the vector between these two centers to guide the DDPM for adaptation to the target domain. However, it suffer from rigid feature translation due to their reliance on fixed geometric directions ($\Delta_{\text{static}}$ in Eq.~1), which fails to capture instance-specific domain shifts. Our key innovation addresses this limitation through a learnable direction adapter that enables dynamic instance-aware adaptation while preserving global domain statistics.

Specifically, given the source dataset S = $\{x_1^S, \dots, x_n^S\}$ and target dataset T = $\{x_1^T, \dots, x_m^T\}$, we extract the image features of each dataset using CLIP \cite{CLIP}. Then, we compute the static cross-domain direction vector $\Delta_{\text{static}}$ from the center of the source domain to the center of the target domain in the feature space:

\begin{equation}
    \Delta_{\text{static}} = \frac{1}{m}\sum_{i=1}^{m}  CLIP(x_{i}^{S})-\frac{1}{n}\sum_{i=1}^{n} CLIP(x_{i}^{T})  
\end{equation}

This captures global feature distribution differences but lacks instance awareness. Our proposed bottleneck network generates batch-specific adjustments conditioned on the source features and the global target domain center for each batch:
\begin{equation}
   \Delta_{\text{dynamic}} = \text{tanh}(W_2 \sigma(W_1(\frac{1}{B}\sum_{i=1}^{B} CLIP(x_{i}^{S}) - \frac{1}{n}\sum_{i=1}^{n} CLIP(x_{i}^{T})))
\end{equation}
Where $W_1 \in \mathbb{R}^{4D \times D}$ and $W_2 \in \mathbb{R}^{D \times 4D}$, $D$ is the dimension of the features, and $B$ is the batch size. 
This allows the model to learn the directional changes for each batch, mapping to a wider range of regions in the target domain. For training stability, we start by using static vectors $\Delta_{\text{static}}$ to guide the model's learning and progressively introduce dynamic adjustments as the learning process advances, enabling broader coverage of the target domain. We employ a learnable gating network to dynamically fuse static and dynamic components:
\begin{equation}
    \Delta = g \odot \Delta_{\text{dynamic}} + (1 - g) \odot \Delta_{\text{static}}
\end{equation}
\begin{equation}
    g(\frac{1}{B}\sum_{i=1}^{B} CLIP(x_{i}^{S})) = \text{sigmoid}(W_g(\sigma(W_e (\frac{1}{B}\sum_{i=1}^{B} CLIP(x_{i}^{S})))))
\end{equation}
Where $W_e \in \mathbb{R}^{512 \times 256}$ and $W_g \in \mathbb{R}^{256 \times 1}$.
We utilize the dynamic directional vector $\Delta$ to constrain the structure of the generated distribution, ensuring it covers the original distribution as much as possible while aligning its center with that of the target distribution. This is achieved through the following distribution consistency loss.

\begin{equation}
    \mathcal{L}_{\text{DC}} = \|CLIP(x^S) + \Delta, CLIP(x_{output}^{S \to T})\|^2
\end{equation}

where $x^S$ is the source image and $x_{output}^{S \to T}$ is the output image in the target domain. Through this loss function, we explicitly enforce the spatial structural consistency between the generated and original distributions during domain adaptation.

Finally, combines diffusion reconstruction \cite{DDPMLoss}, distribution consistency, and style consistency \cite{DDC} losses:
\begin{equation}
    \mathcal{L} = \mathcal{L}_{\text{diff}} + \mathcal{L}_{\text{DC}} + \mathcal{L}_{\text{style}}
\end{equation}

\subsection{Pyramid Hierarchical Alignment}
Medical imaging domains often exhibit discrepancies in intensity distributions while sharing underlying anatomical structures. Traditional domain adaptation methods focusing solely on global feature alignment may fail to capture critical local geometric relationships. As illustrated in Fig.\ref{model}, we propose a adaptation method with pyramid hierarchical feature alignment, addressing both channel-wise semantics and spatial structures. Given paired images $(x^S, x^T)$ from source and target domains, our medical foundation model $f_\theta$ produces multi-scale pyramid features:
\begin{equation}
    \begin{aligned}
        F^{S} &= [f^{(1)}(x^S), \dots, f^{(n)}(x^S)] \\
        F^{T} &= [f^{(1)}(x^T), \dots, f^{(n)}(x^T)]
    \end{aligned}
\end{equation}
where $f^{(k)} \in \mathbb{R}^{B \times C \times H_k \times W_k}$ denotes the $k$-th level feature tensor, $n$ represent the number of features extracted, which we empirically set to 4.

Align features across four pyramid levels to capture organ structures at varying granularities. For each hierarchy level $k$, flatten spatial dimensions while preserving channel correlations and compute cosine similarity between corresponding spatial locations across domains:
    \begin{equation}
        \tilde{f}^{(k)} = \text{reshape}(f^{(k)}, (B, C, H_kW_k))
    \end{equation}

    \begin{equation}
        \mathcal{S}_k = \frac{1}{B(H_kW_k)} \sum_{b=1}^B \sum_{i=1}^{H_kW_k} \frac{ \tilde{f}^{(k)}_b(x^S)[:,i] \cdot \tilde{f}^{(k)}_b(x^T)[:,i] }{ \|\tilde{f}^{(k)}_b(x^S)[:,i]\| \|\tilde{f}^{(k)}_b(x^T)[:,i]\| }
    \end{equation}
    
    Finally, Combine losses across all pyramid levels, enforce position-wise similarity in both channel responses and spatial layouts:
    \begin{equation}
        \mathcal{L}_{\text{align}} = \frac{1}{n} \sum_{k=1}^n (1 - \mathcal{S}_k)
    \end{equation}

We fine-tune the output of each layer using trainable low-rank matrices $\{A_k, B_k\}$ in the attention mechanism:
\begin{equation}
T_i = A_i \times B_i,
\end{equation}
where $A \in \mathbb{R}^{m \times r}$ and $B \in \mathbb{R}^{r \times c}$, $c$ represents the dimensionality of $T_i$, $m$ is the sequence length of $T_i$, and $r$ is the rank with $r \ll c$, which reduces the number of parameters required for fine-tuning. During training, only 0.8\% of the parameters (LoRA matrices) are updated.

The total loss combines segmentation supervision and alignment constraints:
\begin{equation}
\begin{aligned}
    \mathcal{L}_{\text{BCE}}(y, \hat{y}) &= -\left( y \log(\hat{y}) + (1 - y) \log(1 - \hat{y}) \right), \\
    \mathcal{L}_{\text{total}} &= \mathcal{L}_{\text{BCE}}(y, \hat{y}) +  \mathcal{L}_{\text{align}}
    \end{aligned}
\end{equation}

\section{Experiments and Results}
\noindent \textbf{Materials and Evaluation Metrics. }This study utilizes the joint optic cup / optic disc segmentation dataset RIGA+ \cite{FundusDataset1,FundusDataset2,FundusDataset3} and the REFUGE dataset \cite{FundusDataset4}. The RIGA+ dataset provides images from five different domains: BinRushed, Magrabia, BASE1, BASE2, and BASE3. In the REFUGE dataset, the training set and test set were captured using different devices, making them suitable for use as data from different domains. In our experiments, we used REFUGE (Train) as the source domain for model training and REFUGE (Test), BinRushed, and BASE2 as the target domains, which were labeled as I, II, and III, respectively. For image generation tasks, we used IC-LPIPS \cite{ICLPIPS} to measure the diversity of generated images and FID \cite{FID} to evaluate the similarity between the generated images and the target domain. For segmentation tasks, we used the Dice similarity coefficient (D, \%) to assess segmentation performance.

\begin{table}[t]
\centering
\caption{ The evaluation of domain adaptation for the generation and segmentation models in a 10-shot setting is conducted. The DDPM model is trained on the REFUGE (train) dataset. "IC" refers to "IC-LPIPS." For the segmentation task, each method is run three times, and the average and standard deviation are reported. Top three results are highlighted as \colorbox{BDE6CD}{\textbf{best}}, \colorbox{E4EEBC}{second} and \colorbox{FFF8C5}{third}, respectively.}
\label{tab:results}
\resizebox{\columnwidth}{!}{%
\begin{tabular}{c|cccccccc}
\hline \hline
Methods  & \multicolumn{2}{c}{Domain I}                                                                      & \multicolumn{2}{c}{Domain II}                                                                      & \multicolumn{2}{c}{Domain III}                                                                   & \multicolumn{2}{c}{Average}                                                     \\ \hline
Syn      & FID                                         & IC                                    & FID                                          & IC                                    & FID                                         & IC                                   & FID                                    & IC                              \\ \hline
Finetune & 100.83                                      & \cellcolor[HTML]{FFF8C5}0.276                & 209.85                                       & 0.194                                        & \cellcolor[HTML]{FFF8C5}105.28              & \cellcolor[HTML]{E4EEBC}0.308               & 138.65                                 & \cellcolor[HTML]{FFF8C5}0.259         \\
FreezeD \cite{FreezeD} & \cellcolor[HTML]{E4EEBC}84.12               & 0.263                                        & \cellcolor[HTML]{E4EEBC}177.49               & \cellcolor[HTML]{FFF8C5}0.196                & 108.15                                      & 0.268                                       & \cellcolor[HTML]{FFF8C5}123.25         & 0.242                                  \\
DDC \cite{DDC}      & \cellcolor[HTML]{FFF8C5}94.56               & \cellcolor[HTML]{E4EEBC}0.293                & \cellcolor[HTML]{FFF8C5}129.62               & \cellcolor[HTML]{E4EEBC}0.301                & \cellcolor[HTML]{E4EEBC}102.15              & \cellcolor[HTML]{FFF8C5}0.306               & \cellcolor[HTML]{E4EEBC}108.78         & \cellcolor[HTML]{E4EEBC}0.300          \\
ours     & \cellcolor[HTML]{BDE6CD}\textbf{62.79}      & \cellcolor[HTML]{BDE6CD}\textbf{0.371}       & \cellcolor[HTML]{BDE6CD}\textbf{88.27}       & \cellcolor[HTML]{BDE6CD}\textbf{0.347}       & \cellcolor[HTML]{BDE6CD}\textbf{91.52}      & \cellcolor[HTML]{BDE6CD}\textbf{0.345}      & \cellcolor[HTML]{BDE6CD}\textbf{80.86} & \cellcolor[HTML]{BDE6CD}\textbf{0.354} \\ \hline \hline
Seg      & Dice                                        & JI                                           & Dice                                         & JI                                           & Dice                                        & JI                                          & Dice                                     & JI                                   \\ \hline 
Unet     & 78.07$\pm$9.14                                  & 65.28$\pm$11.45                                  & 70.03$\pm$17.61                                  & 57.56$\pm$18.49                                  & 80.62$\pm$9.54                                  & 68.62$\pm$11.54                                 & 76.24                                  & 63.82                                  \\
+Coral \cite{Coral}   & 82.27$\pm$8.34                                  & 70.9$\pm$10.91                                   & 72.37$\pm$14.71                                  & 60.17$\pm$16.24                                  & \cellcolor[HTML]{FFF8C5}84.12$\pm$9.37          & \cellcolor[HTML]{E4EEBC}73.78$\pm$11.4          & 79.58                                  & 68.28                                  \\
+FDA \cite{FDA}     & 81.69$\pm$8.53                                  & 70.73$\pm$11.31                                  & 78.34$\pm$11.32                                  & 67.20$\pm$13.67                                  & \cellcolor[HTML]{BDE6CD}\textbf{85.41$\pm$6.24} & \cellcolor[HTML]{BDE6CD}\textbf{75.30$\pm$8.99} & \cellcolor[HTML]{FFF8C5}81.81          & \cellcolor[HTML]{FFF8C5}71.07          \\
RETFound \cite{zhou2023foundation} & 80.99$\pm$13.34                                 & 71.11$\pm$14.66                                  & 78.71$\pm$15.73                                  & \cellcolor[HTML]{E4EEBC}68.25$\pm$16.87          & 82.27$\pm$10.75                                 & 71.38$\pm$13.12                                 & 80.66                                  & 70.24                                  \\
+Coral \cite{Coral}   & 82.14$\pm$10.27                                 & \cellcolor[HTML]{E4EEBC}74.53$\pm$12.16          & \cellcolor[HTML]{FFF8C5}78.82$\pm$12.28          & 71.31$\pm$7.17                                   & 83.28$\pm$8.59                                  & 72.63$\pm$11.11                                 & 81.41                                  & \cellcolor[HTML]{BDE6CD}\textbf{72.82} \\
+FDA \cite{FDA}    & \cellcolor[HTML]{FFF8C5}84.13$\pm$10.66         & 71.34$\pm$12.94                                  & \cellcolor[HTML]{BDE6CD}\textbf{81.31$\pm$15.30} & \cellcolor[HTML]{FFF8C5}67.87$\pm$17.37          & 83.51$\pm$9.61                                  & 72.56$\pm$12.21                                 & \cellcolor[HTML]{E4EEBC}82.99          & 70.59                                  \\
+Reins \cite{Rein}  & \cellcolor[HTML]{E4EEBC}84.19$\pm$7.21          & \cellcolor[HTML]{FFF8C5}73.92$\pm$9.89           & 77.75$\pm$12.88                                  & 66.48$\pm$15.84                                  & 81.60$\pm$7.37                                  & 70.41$\pm$10.09                                 & 81.18                                  & 70.27                                  \\
+ours    & \cellcolor[HTML]{BDE6CD}\textbf{84.83$\pm$8.09} & \cellcolor[HTML]{BDE6CD}\textbf{74.81$\pm$10.41} & \cellcolor[HTML]{E4EEBC}80.16$\pm$10.87          & \cellcolor[HTML]{BDE6CD}\textbf{68.92$\pm$13.32} & \cellcolor[HTML]{E4EEBC}84.39$\pm$6.20          & \cellcolor[HTML]{FFF8C5}73.53$\pm$8.91          & \cellcolor[HTML]{BDE6CD}\textbf{83.13} & \cellcolor[HTML]{E4EEBC}72.42          \\ \hline \hline
\end{tabular}%
}
\end{table}

\noindent \textbf{Implementation Details.} To ensure consistent resolution between image generation and segmentation, all images are resized to 224×224 pixels. The initial learning rate is set to $1 \times 10^{-3}$, and the maximum number of epochs is fixed at 100 to ensure convergence of all methods. All experiments are implemented using the PyTorch framework and run on five NVIDIA 4090 GPUs. Training DDPM requires 63 GPU hours.

\begin{figure}[!tbp]
	\centering
	\includegraphics[width=\textwidth]{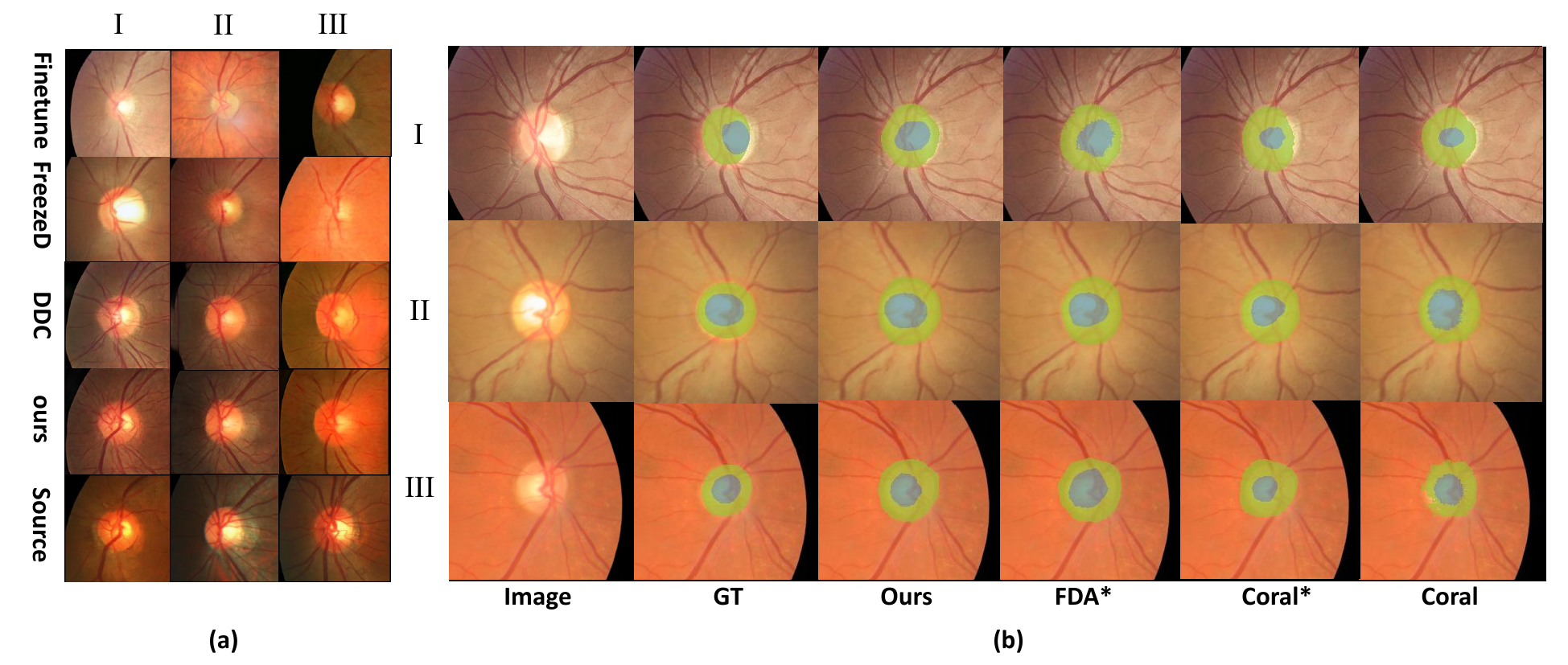}
	\caption{Exemplar results of our method compared to others in the (a) generation and (b) segmentation tasks. Methods marked with “*” are used under MFMs condition.}
	\label{vis}
\end{figure}

\noindent \textbf{Comparative Experiments.} We compared the two stages of MFM-DA with their respective baseline methods and state-of-the-art (SOTA) approaches. In each domain adaptation experiment for image generation models, only 10 images from the target domain were used for training to ensure that the model could access only unlabeled data from a single target domain during the training process. In the feature alignment domain adaptation experiments, the comparison methods were applied not only to the UNet architecture but also to MFMs to enable a more fair comparison. Table \ref{tab:results} presents the results of our method for generation tasks, showing significant improvements over competing approaches. This demonstrates that the Dynamic Instance-Aware Adaptor not only brings the model's distribution closer to that of the target domain, but also enhances diversity, significantly improving the domain adaptation performance. Additionally, the segmentation results in Table \ref{tab:results} reveal that Pyramid Hierarchical Alignment achieves excellent feature alignment for the MFMs. The experimental results further indicate that the MFM-DA fine-tuned MFMs significantly reduces the domain gap with the target domain. Fig. \ref{vis} illustrates some representative results from various methods. Although “Finetune” generates images more similar in style by memorizing target images, our method preserves the structure of the source domain, captures the style of the target domain and produces more diverse images without memorization.

\noindent \textbf{Ablation Analysis.} To evaluate the effectiveness of the multi-level alignment loss used in the medical foundation model, we conducted ablation experiments on four features of the medical foundation model, each with a separate alignment loss. The average performance results are shown in Table \ref{tab:loss}. The results indicate that the use of multi-level alignment loss contributed to performance improvement.

\noindent \textbf{Analysis of Few-shot Setting.} As shown in Table \ref{tab:few-shot}, to evaluate the effectiveness of our method under different few-shot conditions, we tested the cases of 1-shot, 3-shot, 5-shot, 7-shot, and 10-shot with the target domain REFUGE (test). The results indicate that our method remains effective even under extremely few target domain scenarios.

\noindent \textbf{Generalization to different foundation models.} As shown in Table \ref{tab:foundations}, we evaluate the generalizability of our method by applying MFM-DA to various foundation models \cite{CLIP,DINOv2}. The baseline model used is U-Net. The experimental results demonstrate that MFM-DA effectively reduces the domain gap across different foundation models in the medical image segmentation task.

\noindent
\begin{minipage}{\textwidth}
\centering
\begin{minipage}[t]{0.31\textwidth}
\makeatletter\def\@captype{table}
\setlength{\belowcaptionskip}{7pt}
\renewcommand\arraystretch{0.9}
\centering
\setlength\tabcolsep{3pt}
\caption{
Average performance using different levels of loss.
}
\label{tab:loss}
\begin{tabular}{cc}
\hline \hline
Methods & Average \\ \hline
L1      & 84.17   \\
L2      & 84.09   \\
L3      & 84.04   \\
L4      & 83.39   \\
L-All   & 84.83   \\ \hline \hline
\end{tabular}
\end{minipage}%
\hfill%
\begin{minipage}[t]{0.31\textwidth}
\makeatletter\def\@captype{table}
\setlength{\belowcaptionskip}{7pt}
\renewcommand\arraystretch{0.9}
\centering
\setlength\tabcolsep{3pt}
\caption{
Average performance in different few-shot conditions.
}
\label{tab:few-shot}
\begin{tabular}{ccc}
\hline \hline
Setting & LPIPS  & IC    \\ \hline
1-Shot  & 90.17  & -     \\
3-Shot  & 68.93  & 0.356   \\
5-Shot  & 100.47 & 0.315 \\
7-Shot  & 83.74  & 0.335      \\
10-Shot & 62.79  & 0.371 \\ \hline \hline
\end{tabular}%
\end{minipage}
\hfill %
\begin{minipage}[t]{0.31\textwidth}
\makeatletter\def\@captype{table}
\setlength{\belowcaptionskip}{7pt}
\renewcommand\arraystretch{0.9}
\centering
\setlength\tabcolsep{3pt}
\caption{
Average performance using different foundation models.
}
\label{tab:foundations}
\begin{tabular}{cc}
\hline \hline
Methods   & Target \\ \hline
Baseline   & 78.07  \\
DINOv2    & 79.90  \\
+ours     & 80.91  \\
CLIP       & 82.52  \\
+ours       & 83.46  \\ \hline \hline
\end{tabular}%
\end{minipage}
\end{minipage}


\section{Conclusion}
In this paper, we propose a novel few-shot adaptation method for medical foundation models, named MFM-DA. In MFM-DA, the Dynamic Instance-Aware Adaptor generates feature transfer directions for each instance, requiring only 10 images to produce more diverse unlabeled target-domain images. Meanwhile, the Pyramid Hierarchical Alignment loss aligns source-domain and generated target-domain images in the feature space, enabling the medical foundation model to adapt to the target-domain distribution and improve performance on the target domain. Our experimental results on two datasets demonstrate the effectiveness of MFM-DA, showcasing its potential as a promising domain adaptation approach for MFMs. However, MFM-DA has so far only been experimented on fundus images and requires access to source domain data, which raises certain limitations in privacy protection. In our future work, we will extend MFM-DA to more modalities of medical imaging and develop techniques that do not require access to source domain data.

\bibliographystyle{splncs04}
\bibliography{bibliography}

\end{document}